\documentclass[]{bytedance_seed}

\usepackage[toc,page,header]{appendix}
\usepackage{minitoc}
\usepackage{microtype}
\usepackage{graphicx}
\usepackage{subcaption}
\usepackage{booktabs} 
\usepackage{hyperref}
\usepackage{bm}

\usepackage{amsmath}
\usepackage{amssymb}
\usepackage{mathtools}
\usepackage{amsthm}
\theoremstyle{plain}

\theoremstyle{definition}

\theoremstyle{remark}

\usepackage{xspace}
\usepackage{wrapfig}
\usepackage{multirow}   
\usepackage[table,xcdraw]{xcolor} 
\usepackage{tabularx}      
\usepackage{colortbl}

\newcommand{\model}[0]{WoG\xspace}

\title{World Guidance: World Modeling in Condition \\ Space for Action Generation}

\author[1,2]{Yue Su}
\author[2]{Sijin Chen}
\author[1]{Haixin Shi}
\author[1]{Mingyu Liu}
\author[1]{Zhengshen Zhang} 
\author[1]{\\Ningyuan Huang}
\author[1]{Weiheng Zhong}
\author[1]{Zhengbang Zhu}
\author[1, \dagger]{\\Yuxiao Liu}
\author[2, \dagger]{Xihui Liu}

\affiliation[1]{ByteDance Seed}
\affiliation[2]{The University of Hong Kong}

\contribution[\dagger]{Corresponding authors}
\date{\today}
\correspondence{\email{liuyuxiao.876@bytedance.com}, \email{xihuiliu@eee.hku.hk}}
\abstract{
Leveraging future observation modeling to facilitate action generation presents a promising avenue for enhancing the capabilities of Vision-Language-Action (VLA) models. 
However, existing approaches struggle to strike a balance between maintaining efficient, predictable future representations and preserving sufficient fine-grained information to guide precise action generation. 
To address this limitation, we propose \textbf{\model (World Guidance)}, a framework that maps future observations into compact conditions by injecting them into the action inference pipeline. 
The VLA is then trained to simultaneously predict these compressed conditions alongside future actions, thereby achieving effective world modeling within the condition space for action inference. 
We demonstrate that modeling and predicting this condition space not only facilitates fine-grained action generation but also exhibits superior generalization capabilities. Moreover, it learns effectively from substantial human manipulation videos. 
Extensive experiments across both simulation and real-world environments validate that our method significantly outperforms existing methods based on future prediction.
}
\checkdata[Project Page]{\url{https://selen-suyue.github.io/WoGNet/}}

\begin{document}
\maketitle

\abstract{
Leveraging future observation modeling to facilitate action generation presents a promising avenue for enhancing the capabilities of Vision-Language-Action (VLA) models. 
However, existing approaches struggle to strike a balance between maintaining efficient, predictable future representations and preserving sufficient fine-grained information to guide precise action generation. 
To address this limitation, we propose \textbf{\model (World Guidance)}, a framework that maps future observations into compact conditions by injecting them into the action inference pipeline. 
The VLA is then trained to simultaneously predict these compressed conditions alongside future actions, thereby achieving effective world modeling within the condition space for action inference. 
We demonstrate that modeling and predicting this condition space not only facilitates fine-grained action generation but also exhibits superior generalization capabilities. Moreover, it learns effectively from substantial human manipulation videos. 
Extensive experiments across both simulation and real-world environments validate that our method significantly outperforms existing methods based on future prediction.
}
\section{Introduction}
\label{sec:intro}

As systems designed for future action prediction, Vision-Language-Action (VLA) models~\cite{openvla,pi0,pi05} have been expected to improve task performance by developing a more comprehensive ability of modeling the future~\cite{irasim}. Recent works have gone beyond predicting actions alone, investigating whether VLA models can forecast future signals in other modalities and how these predictions can be leveraged to enhance action generation~\citep{geact, worldvla}.

Existing methodologies within this landscape can be broadly categorized into two streams. 
(1) World Action Models~\cite{vpp,worldvla,dreamvla} predict explicit future modalities (such as depth, images, videos), or semantic features from foundation vision models~\cite{dinov2,sam} to facilitate efficient action generation. 
Despite providing rich perceptual cues regarding dynamics, motion, and spatial geometry, prior work~\cite{univla,lamlearn,stamo} indicates that these generic while task-agnostic semantic spaces often contain substantial redundancy for downstream manipulation tasks. This redundancy impedes pretraining efficiency for fine-grained generation and limits cross-scenario scalability, thereby constraining their real-world performance~\cite{univla,hifvla}.
(2) Latent Action Models~\cite{lapa,moto,univla} compress future actions or dynamics into sparse latent representations via reconstruction-based vision supervision, aiming to distill embodiment-agnostic high-level motion patterns. 
While effective for high-level planning and learnable from large-scale video data, these representations have been shown to offer only coarse guidance, lacking the precision required for fine-grained action generation~\cite{lamlearn,motus}.

Collectively, these observations underscore a fundamental trade-off. 
Predicting rich, task-agnostic future representations incurs significant redundancy, thereby increasing computational overhead and hampering performance~\cite{planet,lapo}. 
Conversely, compact latent action spaces typically capture only coarse motion trends, proving insufficient for fine-grained control~\cite{univla,lamlearn}. 
The pivotal challenge, therefore, lies in identifying a predictive space~\cite{jepa} that is both tractable for VLA models to forecast and sufficiently expressive to guide accurate action generation.

\begin{wrapfigure}{r}{0.45\textwidth}
  \centering
  \includegraphics[width=0.45\textwidth]{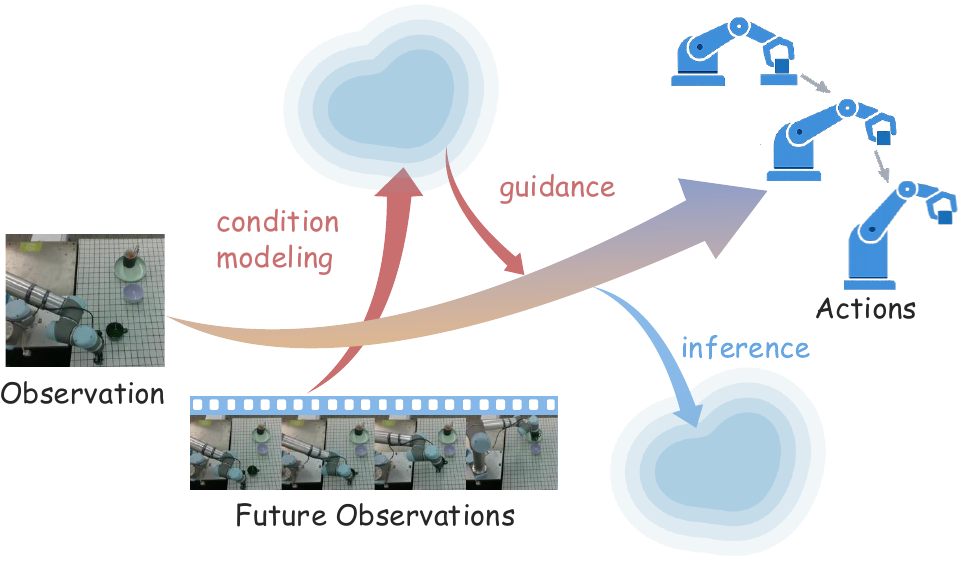}
  \caption{\model first incorporates future observations into the action inference pipeline, projecting them into the condition space for action generation. Subsequently, it decouples future observations from the pipeline and simultaneously predicts these future conditions alongside actions, thereby transferring the knowledge of future conditions into the VLA model.}
  \label{teaser}
\end{wrapfigure}
To address this challenge, we propose \textbf{\model} (World Guidance), a predictive framework that operates in the condition space for action generation. 
We posit that to identify a \textit{non-redundant} predictive space for world action model, the space should satisfy the criterion that its information serves as a sufficient and effective condition for action generation. 
By virtue of this role, such a space is intrinsically highly relevant to action; consequently, for a VLA model inherently designed to model actions, inferring this space becomes a tractable task.
To discover such a space, we argue that an efficient strategy is to directly incorporate future observations as conditions into the action inference pipeline~\cite{flare}. 
The representation encoded through this pipeline thus naturally constitutes the desired efficient condition space.

Specifically, \model follows a two-stage training curriculum. 
In the first stage, we implement the aforementioned design by jointly conditioning action generation on current observations, encoded by a VLM backbone, and future observations derived from frozen foundation vision models, which are queried and compressed by a trainable Q-former based Encoder before being integrated into the action head.
This stage jointly optimizes (i) the encoder to project future observations into this efficient, implicit action-condition space, and (ii) the VLA backbone to leverage these future-conditioned representations for precise action prediction. 
In the second stage, we freeze the Q-former based Encoder to define a stable target space. The VLA is then trained to simultaneously predict this future conditioning representation and the corresponding actions, yielding a model capable of internally anticipating and utilizing future guidance during inference as shown in Figure~\ref{teaser}.

We validate the efficiency of \model in facilitating fine-grained action generation and ensuring robust generalization through extensive simulation and real-world experiments, where it demonstrates substantial improvements over existing methods. 
Furthermore, we show that \model can be effectively improved by learning to model and predict future conditions from large-scale human videos (spanning both action-annotated and unannotated data) or UMI data~\cite{umi}, leading to significant performance gains in real-world robotic deployments.

\section{Related Work}
\label{sec:related}

\subsection{World Action Models}
Leveraging future observation prediction~\cite{flow2act,MBA,g3flow} to extract dynamics~\cite{como,vpp} for robotic manipulation has been extensively studied in imitation learning~\cite{dp,act,dsp,dspv2}.
With the recent advances in VLA models~\cite{openvla,gr3,pi05,pi06}, a growing body of work has begun to exploit the strong reasoning capabilities of Vision-Language Models (VLMs)~\cite{prismatic,paligemma,qwen3} and video generation models~\cite{svd,wan} to perform dynamic modeling, thereby more effectively supporting action inference.

Specifically, certain approaches~\cite{vpp,geact} introduce the intermediate features of video generation models~\cite{svd,cosmos} as world representations, integrating them into action modules to capture future manipulation dynamics. 
Alternative strategies~\cite{upvla,worldvla,rynnvla002} build directly upon VLM backbones, modeling such dynamics through the internal generation of future images~\cite{vqgan}. 
Beyond images generation, the prediction of other explicit modalities, such as depth or optical flow, is also often incorporated into VLA co-training objectives to serve a comparable predictive role~\cite{flowvla,dreamvla}. 
Most recently, studies have also explored directly regressing the latent representations output by foundation vision models, yielding discriminative future features that further refine the precision of action prediction~\cite{dreamvla}. Our method stands out by learning a condition space optimized for action generation, designed to provide more efficient support for enhancing model performance.

\subsection{Latent Action Models}
Latent Action Models (LAMs)~\cite{lapa} have emerged as a response to training VLA models effectively from large-scale, heterogeneous datasets~\cite{oxe}. 
They are built upon the assumption that, despite embodiment variations~\cite{dex-diff}, actions admit high-level representations that are skill-related but embodiment-agnostic. 
Such representations are typically discrete~\cite{vqvae} and are intended to capture coarse motion trends, thereby facilitating high-level planning.

Mainstream LAMs~\cite{lapa,moto,univla} compress heterogeneous actions across embodiments by using visual reconstruction objectives, after which a VLA model first predicts discretized latent actions and subsequently decodes them into fine-grained actions suitable for downstream tasks~\cite{go1}. 
These reconstructions may rely on either generative~\cite{lapa} or discriminative representations~\cite{univla}, but are all designed to produce compact and accurate latent action spaces. 

However, prior studies~\cite{lamlearn,clap} have observed that the compression performed by LAMs often resembles PCA-like~\cite{pca} extraction of maximum-variance signals, yielding coarse planning representations and, in some cases, introducing noise from scenarios that is weakly correlated with actions. 
To mitigate these limitations, recent work has incorporated action reconstruction into the generative objectives of LAMs to strengthen the mapping between latent actions and control~\cite{vita,xr1,clap}. 
Other approaches further introduce video generation as an auxiliary co-training objective in the LAM framework, aiming to compensate for the lack of fine-grained action guidance~\cite{vita,vipra}. While sharing the goal of providing rich guidance, we posit that introducing condition prediction instead of video reconstruction in latent space is efficient.

\section{Method}
\label{sec:method}

\subsection{Problem Formulation}
We consider the problem of predicting future $T$ actions $\bm{A}_{t:t+T}$ given the current observation $\bm{O}_t$ at time step $t$, and a language instruction $\bm{l}$.
The VLM backbone encodes the observation and instruction as $\bm{z} \leftarrow f(\bm{O}_t, \bm{l})$, which is then fed into an action head to generate actions by maximizing the likelihood of $\bm{P}(\bm{A}_{t:t+T}\mid\bm{z})$.

In the first stage of training \model, observations from the next $T$ future time steps are guided to be compressed into a condition space, denoted as $\bm{O}^{c}_{t:t+T}$.
Together with the current latent representation $\bm{z}$, the compressed future condition is fed into the action head as a guidance to generate actions by modeling
$
   \bm{P}(\bm{A}_{t:t+T}\mid\bm{z},\, \bm{O}^{c}_{t:t+T}). 
$

Through this process, the VLA model learns to encode conditions from the current observation while leveraging complementary conditions derived from future observations for action prediction, and simultaneously acquires a compact representation of future observations.

\begin{figure*}[!t]
  \centering
    \includegraphics[width=\textwidth]{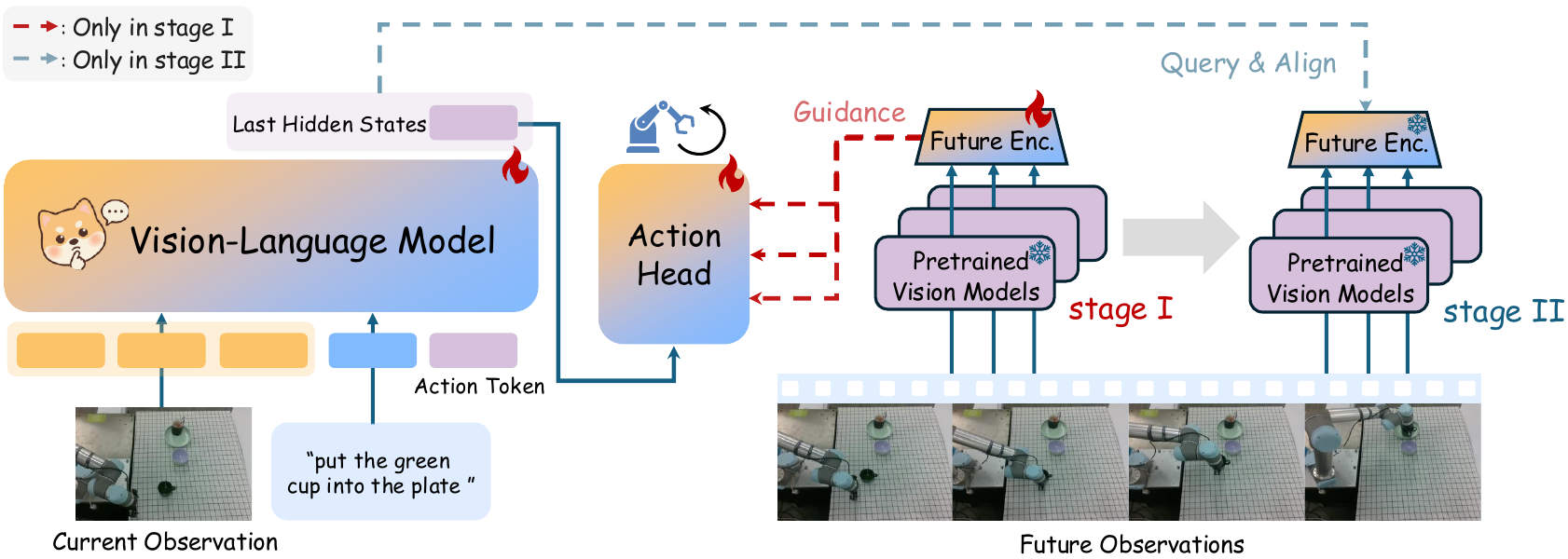}
    \caption{\textbf{Overview of \model}. \model is trained in two stages.
In the first stage, future observations encoded by frozen vision foundation models are queried and compressed by a trainable Q-former-based Future Encoder to form condition representations, which, together with VLM-encoded current observations and instructions, are used for action prediction.
In the second stage, the encoder and vision models are frozen, and the VLM backbone is trained to align with the conditions while predicting actions.}
\label{pipeline}
\end{figure*}

However, given our goal of performing action inference solely based on the current observation at test time, the subset of conditions derived from the future is not accessible and must be inferred from the current observation. Under the assumption of deterministic environmental dynamics, the complete inference process should be formulated as:
\begin{equation}
\bm{P}(\bm{A}_{t:t+T}, \bm{O}^{c}_{t:t+T} \mid \bm{z})
  = \bm{P}(\bm{A}_{t:t+T} \mid \bm{z},\, \bm{O}^{c}_{t:t+T}) \bm{P}(\bm{O}^{c}_{t:t+T} \mid \bm{z}).
\end{equation}
Therefore, in the second stage, \model is trained with two objectives.
At the output of the VLM, supervision is applied to predict the future condition by modeling $\bm{P}(\bm{O}^{c}_{t:t+T} \mid \bm{z})$.
Then at the output of the action head, supervision is imposed on action prediction as $\bm{P}(\bm{A}_{t:t+T}\mid\bm{z})$, by optimizing this marginal action likelihood as the joint distribution under the deterministic coupling of dynamics.
Together, these objectives transfer the knowledge of future condition into the VLM backbone itself. The overall process of \model is shown in Figure~\ref{pipeline}.

\subsection{Stage I: World Guidance}
In the first stage, we use a Prismatic VLM~\cite{prismatic} adopted in OpenVLA~\cite{openvla} as the VLM backbone.
The current observation and language instruction are encoded by the VLM backbone to obtain the latent representation $\bm{z}$ (Following~\cite{cogact}, we represent it as the output feature of the last learnable token), which is then fed into a DiT~\cite{DiT} action head for action generation.

For future observations, they are first encoded using a combination of frozen pretrained vision models to obtain high-level representations.
By default, we extract discriminant and semantic features by DINOv2~\cite{dinov2} and generative features by Wan VAE Encoder~\cite{wan}.
Such a combination of foundation vision models is extensible and can be replaced by other pretrained visual encoders~\cite{sam,siglip}.
After being projected to a unified embedding dimension, these future dynamic features are processed by a learnable Q-Former-based Encoder~\cite{blip2}, which queries action-relevant features and projects them into low-dimensional conditioning representations as $\bm{O}^{c}$.

The queried representations $\bm{O}^{c}$ are then injected into each DiT block, where they perform cross-attention with $\bm{z}$ to enable action prediction conditioned on both the current and future information.
We adopt rectified flow~\cite{rectified_flow} to predict the velocity field:
\begin{equation}
  \mathcal{L}_{\text{I}} =
\mathbb{E}_{\tau,\, A}
\Bigl[
\bigl\|
v_{\theta}(A_\tau, \tau , \bm{z},\, \bm{O}^{c})
-
v^{\ast}
\bigr\|_2^2
\Bigr],
\end{equation}

where $\tau \in [0,1]$ denotes the scheduling timestep,
$v_{\theta}$ and $v^{\ast}$ denote the predicted and target velocity, respectively.

\subsection{Stage II: World Inference}
In the second stage, the Q-Former and the projectors used for encoding future observations are frozen.
\model is designed to jointly train action prediction $\bm{P}(\bm{A}_{t:t+T}\mid\bm{z})$ and future condition prediction $\bm{P}(\bm{O}^c_{t:t+T}\mid\bm{z})$.

Specifically, we introduce learnable query embeddings to attend to the last hidden states of the VLM output with a cross-attention paradigm, and align the queried representations with the frozen future condition representations $\bm{O}^{c}$ produced by the Encoder.
Afterwards, only the VLM output $\bm{z}$ is fed into the DiT head as input to predict actions. The loss is formulated as:
\begin{equation}
  \mathcal{L}_{\text{II}} =
\mathbb{E}_{\tau,\, A}
\Bigl[
\bigl\|
v_{\theta}(A_\tau, \tau , \bm{z})
-
v^{\ast}
\bigr\|_2^2
\Bigr]+1-\mathcal{S}\Bigl[\bm{O}^c , f_q(\bm{O},\bm{l}) \Bigr],
\end{equation}

where $f_q(\bm{O}, \bm{l})$ denotes the queried last hidden state of the VLM, and $\mathcal{S}[\cdot,\cdot]$ denotes the cosine similarity.

In this stage, the future condition is decoupled from the action head and becomes one of the prediction targets of the VLM.
Through supervision, the VLM is encouraged to encode future condition information in its internal representations, enabling the model to perform complete action inference solely based on $\bm{z}$.
As a result, \model is transformed into a \textit{self-guided} model. More details about the query mechanism of \model are provided in Appendix~\ref{app:model}.

\subsection{Learning From Human Manipulation}
\label{sec:humanumi}
Our method is also easy to extend to learn from human manipulation videos, which can be incorporated in two complementary ways.

\textbf{(1)}
A small amount of human videos with action annotations is introduced in the first stage, together with robot data, to expand the condition space and capture manipulation knowledge absent from robot demonstrations.
In the second stage, a much larger collection of unlabeled human videos is incorporated to supervise future condition prediction, while action supervision is applied only to robot data and optionally to the annotated human subset.
This strategy enables the model to learn more sufficient and generalizable future conditions from large-scale video data.

\textbf{(2)}
No action-annotated human data is required.
Human videos are directly introduced in the second stage to supervise condition prediction, while the action prediction branch of human videos is masked.
This setting assumes that the first stage trained on robot data already learns a sufficiently expressive condition space, and that many of these conditions, such as object motion dynamics, are shared with human manipulation videos.
Under this assumption, second-stage training alone can further improve the model’s ability to predict future conditions and enhance generalization across diverse scenarios.

We provide comprehensive validation of both strategies in Section~\ref{sec:human_exp}, demonstrating their effectiveness.

In addition, UMI can also be regarded as a universal interface for robot data.
We expect that training on UMI data can further benefit our study, as it provides a means to evaluate whether the condition space of \model can be stably modeled under egocentric observations and unseen embodiments, and subsequently transferred to our target workspace.
We conduct corresponding validation in Section~\ref{sec:umi_exp}.

\section{Simulation Experiments}
\label{sec:sim}
\subsection{Setup}
\textbf{Evaluation Setup.} We evaluate our method in the SIMPLER simulation environment~\cite{simplerenv}, which includes two robotic configurations: the Google Robot and WidowX.
During evaluation, the model operates in a closed-loop manner and receives only a single RGB observation at each time step. Detailed training and evaluation settings of our method can be found in Appendix~\ref{app:sim}.

\textbf{Baselines.} In the baseline comparison, we aim to highlight both the advantages of \model over conventional VLA approaches and its distinctions from closely related methods.
Accordingly, we include a broad and representative set of VLA baselines spanning multiple paradigms. Specifically, our comparisons cover:

(i) Conventional VLA methods that directly map visual-language observations to actions: $\pi_0$~\cite{pi0}, $\pi_0$-FAST~\cite{fast}, OpenVLA~\cite{openvla} and GR00T-N1~\cite{gr00t}; 

(ii) Latent Action Models: Moto~\cite{moto} and UniVLA~\cite{univla};

(iii) World Action Models that leverage video prediction to capture dynamics: DeFI~\cite{defi};

(iv) Methods that perform latent action modeling with future video generation: VITA~\cite{vita} and ViPRA~\cite{vipra}.

\begin{table*}[t]
\centering
\caption{\textbf{SimplerEnv evaluation across different models on Google Robot tasks.} Mv Near: Move Near, Drawer: Open/Close Drawer. \model outperforms existing methods across the majority of tasks, with particularly notable gains in scenarios that necessitate efficient trajectory planning and collision avoidance.}
\label{tab:simplerenv}
\resizebox{0.75\textwidth}{!}{%
\begin{tabular}{l|cccc|cccc|c}
\toprule
\multirow{2}{*}{\textbf{Model}} & \multicolumn{4}{c|}{\textbf{Visual Matching}} & \multicolumn{4}{c|}{\textbf{Variant Aggregation}} & \textbf{Overall} \\
 & Pick Coke & Mv Near & Drawer & Avg. & Pick Coke & Mv Near & Drawer & Avg. & Avg. \\ 
\midrule
$\pi_0$~\cite{pi0}  & 72.7\% & 65.3\% & 38.3\% & 58.8\% & 75.2\% & 63.7\% & 25.6\% & 54.8\% & 56.8\% \\
$\pi_0$-FAST~\cite{fast}  & 75.3\% & 67.5\% & 42.9\% & 61.9\% & 77.6\% & 68.2\% & \textbf{31.3\%} & 59.0\% & 60.5\% \\
OpenVLA~\cite{openvla}  & 16.3\% & 46.2\% & 35.6\% & 32.7\% & 54.5\% & 47.7\% & 17.7\% & 39.8\% & 33.8\% \\
GR00T-N1~\cite{gr00t}  & 47.0\% & 70.0\% & 18.1\% & 45.0\% & 78.8\% & 62.5\% & 13.2\% & 51.5\% & 48.4\% \\
Moto~\cite{moto}  & 74.0\% & 60.4\% & 43.1\% & 59.2\% & -- & -- & -- & -- & -- \\
VITA~\cite{vita}  & 57.5\% & 55.8\% & 58.9\% & 57.4\% & -- & -- & -- & -- & -- \\
DeFI~\cite{defi} & 54.2\% & 60.7\% & 38.6\% & 51.2\% & 53.9\% & 58.2\% & 24.0\% & 45.4\% & 48.3\% \\
\rowcolor[HTML]{EFEFEF} 
\textbf{\model} & \textbf{89.0\%} & \textbf{82.5\%} & \textbf{62.5\%} & \textbf{78.0\%} & \textbf{87.9\%} & \textbf{75.0\%} & 19.3\% & \textbf{60.7\%} & \textbf{69.4\%} \\ 
\bottomrule
\end{tabular}%
}
\end{table*}
\begin{table*}[t]
\centering
\caption{\textbf{SimplerEnv evaluation across different models on  WidowX Robot tasks.} \model outperforms existing methods across a broad spectrum of Pick-and-Place tasks, notably surpassing approaches that jointly model video prediction and latent actions.}
\label{tab:simplerenv-widow}
\resizebox{\textwidth}{!}{%
\begin{tabular}{l|cc|cc|cc|cc|cc}
\toprule
\multirow{2}{*}{\textbf{Model}} & \multicolumn{2}{c|}{\textbf{Put Spoon on Towel}} & \multicolumn{2}{c|}{\textbf{Stack Green on Yellow}} & \multicolumn{2}{c|}{\textbf{Put Carrot on Plate}} & \multicolumn{2}{c|}{\textbf{Put Eggplant in Basket}} & \multicolumn{2}{c}{\textbf{Overall Average}} \\
 & Grasp Spoon & Success & Grasp G Block & Success & Grasp Carrot & Success & Grasp Eggplant & Success & Grasp Avg. & Success Avg. \\ 
\midrule
$\pi_0$~\cite{pi0} & 45.8\% & 29.1\% & 25.0\% & 0.0\% & 50.0\% & 16.7\% & 91.6\% & 62.5\% & 40.1\% & 27.1\%\\
$\pi_0$-FAST~\cite{fast}  & 62.5\% & 29.1\% & 58.5\% & 21.9\% & 54.0\% & 10.8\% & 83.3\% & 66.6\% & 48.3\% & 32.1\%\\
OpenVLA~\cite{openvla}  & 4.1\% & 0.0\% & 33.0\% & 0.0\% & 12.5\% & 0.0\% & 8.3\% & 4.1\% & 7.8\% & 1.1\% \\
GR00T-N1~\cite{gr00t}  & 83.3\% & 62.5\% & 54.2\% & 45.8\% & 70.8\% & 16.7\% & 41.7\% & 20.8\% & 49.5\% & 36.5\%\\
UniVLA~\cite{univla} & 76.4\%  & 52.8\% & 66.7\% & 2.8\% & \textbf{79.2\%} & \textbf{55.6\%} & 87.5\% & 66.7\% & 77.5\% & 45.6\% \\
ViPRA~\cite{vipra}  & 79.2\% & 66.7\% & 62.5\%  & \textbf{54.2\%} & 54.2\% & 50.0\% & 91.7\% & 79.2\% & 71.9\% & 62.5\% \\
\rowcolor[HTML]{EFEFEF} 
\textbf{\model} & \textbf{95.8\%} & \textbf{79.2\%} & \textbf{75.0\%} & 33.0\% & 70.8\% & 50.0\%  & \textbf{100.0\%} & \textbf{91.7\%} & \textbf{85.4\%} & \textbf{63.5\%} \\ 
\bottomrule
\end{tabular}%
}
\end{table*}
\subsection{Results}
Most tasks in the SIMPLER benchmark belong to the pick-and-place type, where successful execution critically depends on both dynamic trajectory planning and precise end-effector pose prediction.
In particular, obstacle avoidance during motion requires anticipating scene dynamics, while grasping and placement demand accurate reasoning about future contact and collision constraints.

As shown in Table~\ref{tab:simplerenv} and Table~\ref{tab:simplerenv-widow}, \model achieves strong and consistent performance improvements over all baselines across the majority of tasks.
In scenes where interfering objects are present, such as \textit{Move Near}, our approach exhibits markedly superior trajectory planning behavior, effectively navigating dynamic interference and maintaining stable execution.
This highlights the benefit of incorporating future-aware conditions for motion reasoning under complex environmental dynamics.

Furthermore, in broad P\&P tasks like \textit{Pick Coke} and \textit{Put Spoon}, \model substantially improves the accuracy of future grasp and placement pose prediction.
By extracting future semantics in a manner that is both sufficient and compact, our approach achieves superior accuracy in estimating target poses, thereby enhancing positioning precision and collision avoidance capabilities.
It is particularly noteworthy that, while our paradigm shares the similar underlying set of future observations with methods combining latent actions with future video generation, it diverges fundamentally in how this information is utilized.
Unlike full-scale video prediction, which aims to reconstruct all visual information, our method selectively extracts critical semantics into the condition space.
This strategy yields a more robust representation that effectively mitigates the propagation of visual prediction errors into the action space, ultimately facilitating action prediction with significantly higher precision.

We also note that in a small subset of tasks characterized by action constraints, such as \textit{Stack Green on Yellow} and \textit{Drawer} for their requirement on accurate relative position of stacks or gripper and drawer, the performance gains are comparatively smaller.
This is primarily attributable to the limited spatial resolution of the current backbone and the inherent difficulty of modeling fine-grained geometry~\cite{falcon,internvlam1} through current dynamic prediction alone. 

\subsection{Pretrained Encoder Configuration}
\label{pec}
To assess the adaptability of our framework, we instantiate \model with 3 different combinations of foundation vision encoders: (i) DINOv2~\cite{dinov2} only, (ii) DINOv2 paired with SigLIP~\cite{siglip}, and (iii) DINOv2 paired with the Wan VAE encoder~\cite{wan}. The results of these configurations are presented in Table~\ref{app-rt1} and Table~\ref{app-bri}.

\begin{table*}[h!]
\centering
\caption{\textbf{SimplerEnv evaluation across different pretrained encoder configurations on Google Robot tasks.} Mv Near: Move Near, Drawer: Open/Close Drawer.}
\label{app-rt1}
\resizebox{0.8\textwidth}{!}{%
\begin{tabular}{l|cccc|cccc|c}
\toprule
\multirow{2}{*}{\textbf{Model}} & \multicolumn{4}{c|}{\textbf{Visual Matching}} & \multicolumn{4}{c|}{\textbf{Variant Aggregation}} & \textbf{Overall} \\
 & Pick Coke & Mv Near & Drawer & Avg. & Pick Coke & Mv Near & Drawer & Avg. & Avg. \\ 
\midrule
\model(dino) & 96.0\% & 85.8\% & 56.0\% & 79.3\% & 88.1\% & 81.0\% & 9.8\% & 59.6\% & 69.5\% \\
\model(dino-siglip) & 89.0\% & 82.5\% & \textbf{62.5\%} & 78.0\% & 87.9\% & 75.0\% & \textbf{19.3\%} & \textbf{60.7\%} & 69.4\% \\ 
\model(dino-vae) & \textbf{97.7\%} & \textbf{87.1\%} & 61.1\% & \textbf{82.0\%} & \textbf{89.7\%} & \textbf{77.5\%} & 12.4\% & 59.9\% & \textbf{70.9\%} \\
\bottomrule
\end{tabular}%
}
\end{table*}

\begin{table*}[h!]
\centering
\caption{\textbf{SimplerEnv evaluation across different different pretrained encoder configurations on  WidowX Robot tasks.}}
\label{app-bri}
\resizebox{\textwidth}{!}{%
\begin{tabular}{l|cc|cc|cc|cc|cc}
\toprule
\multirow{2}{*}{\textbf{Model}} & \multicolumn{2}{c|}{\textbf{Put Spoon on Towel}} & \multicolumn{2}{c|}{\textbf{Stack Green on Yellow}} & \multicolumn{2}{c|}{\textbf{Put Carrot on Plate}} & \multicolumn{2}{c|}{\textbf{Put Eggplant in Basket}} & \multicolumn{2}{c}{\textbf{Overall Average}} \\
 & Grasp Spoon & Success & Grasp G Block & Success & Grasp Carrot & Success & Grasp Eggplant & Success & Grasp Avg. & Success Avg. \\ 
\midrule
\model(dino) & 83.3\% & 66.7\% & 41.7\% & 8.3\% & 50.0\% & 29.2\%  & \textbf{100.0\%} & 91.7\% & 68.8\% & 49.0\% \\
\model(dino-siglip) & \textbf{95.8\%} & \textbf{79.2\%} & \textbf{75.0\%} & \textbf{33.0\%} & 70.8\% & \textbf{50.0\%}  & \textbf{100.0\%} & 91.7\% & 85.4\% & \textbf{63.5\%} \\
\model(dino-vae) & \textbf{95.8\%} & 54.2\% & 66.7\% & 29.2\% & \textbf{83.3\%} & \textbf{50.0\%}  & \textbf{100.0\%} & \textbf{100.0\%} & \textbf{86.4\%} & 58.4\% \\ 
\bottomrule
\end{tabular}%
}
\end{table*}
The quantitative results presented in Table~\ref{app-rt1} and Table~\ref{app-bri} reveal three key insights regarding the design of the condition space:

\begin{itemize}
    \item \textbf{Benefits of Enriched Representations:} Both the SigLIP and Wan VAE augmented configurations consistently outperform the distinct DINOv2 baseline. This validates that enriching the condition space with high-level semantics or temporal dynamics significantly enhances policy robustness.
    
    \item \textbf{VAE for Trajectory Planning:} The \textit{dino-vae} variant demonstrates superior performance on Google Robot tasks (e.g., \textit{Pick Coke}, \textit{Move Near}), achieving the highest overall success rate of 70.9\%. We attribute this to the VAE encoder's ability to compress spatiotemporal information, which effectively aids the policy in modeling object dynamics and planning smooth trajectories under environmental disturbances.
    
    \item \textbf{SigLIP for Spatial Precision:} On tasks that demand fine-grained spatial reasoning, such as \textit{Stack Green on Yellow}, the \textit{dino-siglip} variant exhibits a distinct advantage, surpassing \textit{dino-vae} by a notable margin in success rate (33.0\% vs. 29.2\%). This suggests that explicit high-level semantic alignment provided by SigLIP is critical for handling tasks with spatial constraints and precise positioning requirements.
\end{itemize}

It's noteworthy that the integration of SigLIP enriches the condition space with high-level semantics, partially mitigating spatial precision deficits and ensuring robust performance across the SIMPLER benchmark. Thus, we utilize the \textit{dino-siglip} configuration as the default for simulation. Nevertheless, we contend that modeling fine-grained spatial constraints remains a persistent challenge independent of the future encoder. Solving this requires dedicated spatial mechanisms~\cite{vggt,streamvggt} or historical observation modeling, which lies beyond the scope of this work. Since our real-world experiments are tailored to validate trajectory planning, we retain the VAE-incorporated framework in physical evaluations to fully exploit its superior capacity.

\subsection{Ablation of Future Encoder}
\label{acm}

To verify the effectiveness of utilizing the Future Encoder to query compact condition representations from vision foundation models, we conduct an ablation study focusing on the encoder mechanism under the \textit{dino-vae} configuration. Specifically, we evaluate the following variants to assess the necessity of the Future Encoder:
\begin{itemize}
\item \textbf{\model w/o Future Enc.:} This variant completely excludes the Future Encoder from both training stages. Instead of compressing visual features, we instantiate a set of learnable embeddings equivalent in number to the total tokens of the DINOv2 and VAE feature maps. These embeddings query the last hidden states of the VLM and are supervised to align directly with the full, uncompressed feature maps from the frozen vision models in the total 150k training steps.
\item \textbf{\model w/o Future Enc. in Stage-II:} In this setting, the Future Encoder is retained during the first stage to inject compact guidance into the action generation pipeline. However, in the second stage, the encoder is discarded, and the VLM is trained to align its predictive representations with the full, uncompressed feature maps of the foundation models, identical to the alignment strategy in \textit{\model w/o Future Enc.}
\item \textbf{\model w. Future Enc.:} The standard training paradigm of our method, where the VLM is supervised to align with the low-dimensional condition representations $\boldsymbol{O}^c$ queried and compressed by the Future Encoder.
\end{itemize}

\begin{table*}[h!]
\centering
\caption{\textbf{SimplerEnv evaluation across different Future Encoder ablation variants on Google Robot tasks.} Mv Near: Move Near, Drawer: Open/Close Drawer.}
\label{app-simabl-rt1}
\resizebox{0.8\textwidth}{!}{%
\begin{tabular}{l|cccc|cccc|c}
\toprule
\multirow{2}{*}{\textbf{Model}} & \multicolumn{4}{c|}{\textbf{Visual Matching}} & \multicolumn{4}{c|}{\textbf{Variant Aggregation}} & \textbf{Overall} \\
 & Pick Coke & Mv Near & Drawer & Avg. & Pick Coke & Mv Near & Drawer & Avg. & Avg. \\ 
\midrule
\model w/o Future Enc. & 93.3\% & 79.2\% & 52.8\% & 75.1\% & 87.3\% & 72.9\% & \textbf{14.8\%} & 58.3\% & 66.7\% \\
\model w/o Future Enc. in Stage-II & 91.3\% & \textbf{87.1\%} & 46.3\% & 74.9\% & 88.0\% & \textbf{77.5\%} & 9.8\% & 58.4\% & 66.7\% \\
\model w. Future Enc. & \textbf{97.7\%} & \textbf{87.1\%} & \textbf{61.1\%} & \textbf{82.0\%} & \textbf{89.7\%} & \textbf{77.5\%} & 12.4\% & \textbf{59.9\%} & \textbf{70.9\%} \\
\bottomrule
\end{tabular}%
}
\end{table*}

\begin{table*}[h!]
\centering
\caption{\textbf{SimplerEnv evaluation across different Future Encoder ablation variants on  WidowX Robot tasks.}}
\label{app-simabl-bri}
\resizebox{\textwidth}{!}{%
\begin{tabular}{l|cc|cc|cc|cc|cc}
\toprule
\multirow{2}{*}{\textbf{Model}} & \multicolumn{2}{c|}{\textbf{Put Spoon on Towel}} & \multicolumn{2}{c|}{\textbf{Stack Green on Yellow}} & \multicolumn{2}{c|}{\textbf{Put Carrot on Plate}} & \multicolumn{2}{c|}{\textbf{Put Eggplant in Basket}} & \multicolumn{2}{c}{\textbf{Overall Average}} \\
 & Grasp Spoon & Success & Grasp G Block & Success & Grasp Carrot & Success & Grasp Eggplant & Success & Grasp Avg. & Success Avg. \\ 
\midrule
\model w/o Future Enc. & 75.0\% & 54.2\% & 58.3\% & \textbf{29.2\%} & 75.0\% & \textbf{54.2\%}  & 91.7\% & 91.7\% & 75.0\% & 57.3\% \\
\model w/o Future Enc. in Stage-II & 83.3\% & \textbf{70.8\%} & 58.3\% & \textbf{29.2\%} & 50.0\% & 37.5\%  & 95.8\% & 91.7\% & 71.8\% & 57.3\% \\ 
\model w. Future Enc. & \textbf{95.8\%} & 54.2\% & \textbf{66.7\%} & \textbf{29.2\%} & \textbf{83.3\%} & 50.0\%  & \textbf{100.0\%} & \textbf{100.0\%} & \textbf{86.4\%} & \textbf{58.4\%} \\ 
\bottomrule
\end{tabular}%
}
\end{table*}

The experimental results, detailed in Table~\ref{app-simabl-rt1} and Table~\ref{app-simabl-bri}, demonstrate that the proposed \model, which utilizes the Future Encoder to query conditions, consistently outperforms variants that lack this component across the majority of tasks. This advantage is particularly pronounced in the success rates of grasping and picking phases, suggesting that the condition extraction mechanism effectively leverages the potential of the underlying vision foundation models and keep generalizable during various setups, by amplifying the trajectory planning capabilities inherent to the \textit{dino-vae} configuration during action execution.

We also observe that the method does not yield a distinct advantage in spatially sensitive placement tasks, such as \textit{Stack}. This reveals a limitation of the current paradigm: while \model effectively magnifies the inherent strengths of foundation models for action generation, it cannot inject the additional semantic information required to compensate for their intrinsic deficiencies in specific capabilities.

\section{Real-World Experiments}
\label{sec:real}
\begin{figure*}[!t]
  \centering
    \includegraphics[width=\textwidth]{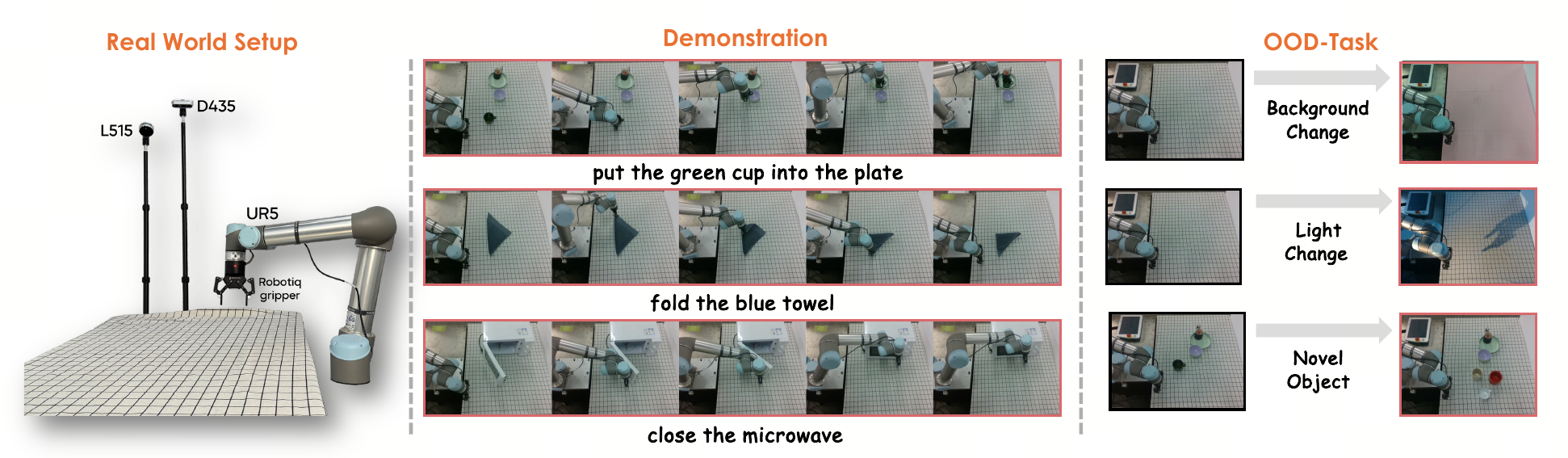}
    \caption{\textbf{Overview of our real-world experiment setup}. The figure shows our robotic platform and sensors (left), the execution of the three tasks under the in-distribution setup (middle), and the modifications applied for the out-of-distribution setup (right).
}
    \label{realexp}
\end{figure*}
\subsection{Setup}
\textbf{In-Distribution (ID) Setup.}
We design the following 3 tasks for evaluation and collect corresponding expert demonstrations, forming our In-Distribution (ID) setup:

\textit{Pick and Place} (P\&P) is a rigid-body manipulation, where the robot picks and places a green cup into a plate following the given instruction.
Successful execution requires avoiding table-top obstacles and collisions with objects already in the plate, which evaluates the model’s ability to predict action-relevant dynamics for collision-aware trajectory planning.
We collect 100 expert demonstrations for this task.

\textit{Close the Microwave} is an articulated-object manipulation task, where the robot is required to close the microwave door.
This task evaluates the policy’s ability to predict and control articulated rotational dynamics. We collect 100 expert demonstrations for this task.

\textit{Fold the Towel} is a deformable manipulation task, where the robot grasps one bottom corner of a towel folded into an isosceles triangle and aligns it with the other bottom corner.
The task is considered successful if the distance between the two corners is less than $5\,\mathrm{cm}$.
We collect 200 expert demonstrations for this task.

\textbf{Out-of-Distribution (OOD) Setup.}
To evaluate generalization, we construct 3 conditions unseen in the expert demonstrations, forming the Out-of-Distribution (OOD) setup:

\textit{Background Change}: Expert demonstrations are collected using a single tablecloth, while evaluation is conducted with a different tablecloth to introduce background variation.

\textit{Light Change}: During evaluation, a numerically controlled light source with fixed intensity is projected onto the workspace from a fixed position, introducing lighting conditions not covered by the expert demonstrations.

\textit{Novel Object}: For \textit{Fold the Towel}, evaluation is conducted using a unseen towel. 
For \textit{Pick and Place}, we replace the original cup with cups of different colors and shapes and modify the corresponding instructions accordingly.

\textbf{Platform.}
We use a UR5 robotic arm equipped with a Robotiq 2F-85 gripper for manipulation.
A top-down Intel RealSense D435 and an L515 camera are mounted to provide global observations, while only the D435 is used in this work.
All devices are connected to a workstation with an NVIDIA RTX 4090 GPU for model inference and control.

\textbf{Baselines.} 
We select UniVLA~\cite{univla}, a Latent Action Model, and VPP~\cite{vpp}, which leverages future video prediction, both of which have been widely validated in real-world robotic settings, as baselines.
In addition, we compare against variants that ablate each training stage, as well as models trained on different data sources, to comprehensively assess the contribution of our approach.

\textbf{Protocols.} 
For the real-world evaluation, 20 trials are performed per method for each task. All methods are evaluated under closely matched randomized initial scene configurations for each trial. 

The above experiment setups are shown in the Figure~\ref{realexp}. Detailed training and evaluation settings of our method can be found in Appendix~\ref{app:real}.

\subsection{Effective and Generalizable Manipulation}
The performance results of \model and baselines are summarized in Table~\ref{tab:method_comp}.
Through these evaluations, we aim to assess whether \model can effectively model future condition and leverage its prediction to facilitate action generation in dynamic interactions.
We further examine whether incorporating future observation prediction leads to overfitting to ID scenarios, potentially degrading generalization performance under OOD setup.

\begin{table*}[h!]
    \centering
    \caption{\textbf{Detailed performance comparison across all tasks.} We report the success rate (\%) for In-Distribution (ID) and various Out-of-Distribution scenarios. For OOD settings, scores are formatted as \textit{ID $\to$ OOD}. Best results are highlighted in \textbf{bold}.}
    \label{tab:method_comp}
    \definecolor{tabgray}{gray}{0.95}
    \setlength{\tabcolsep}{4pt} 
    \renewcommand{\arraystretch}{1.12} 
    \resizebox{0.9\textwidth}{!}{%
        \begin{tabular}{l | c | ccc | cccc}
            \toprule
            \multirow{2}{*}{\textbf{Model}} & \textbf{Microwave} & \multicolumn{3}{c|}{\textbf{Pick and Place}} & \multicolumn{4}{c}{\textbf{Fold the Towel}} \\
            \cmidrule(lr){2-2} \cmidrule(lr){3-5} \cmidrule(lr){6-9}
             & ID & ID & Background & Novel Object & ID & Background & Light Change & Novel Object \\
            \midrule
            
            UniVLA~\cite{univla} & 80\% 
                   & 25\% & 25\%$\to$20\% & 25\%$\to$10\% 
                   & 20\% & 20\%$\to$20\% & 20\%$\to$10\% & 20\%$\to$10\% \\
            
            VPP~\cite{vpp}    & 90\% 
                   & 55\% & 55\%$\to$30\% & 55\%$\to$15\% 
                   & 45\% & 45\%$\to$30\% & 45\%$\to$20\% & 45\%$\to$30\% \\
            
            \rowcolor{tabgray} 
            \textbf{\model} & \textbf{100\%} 
                            & \textbf{60\%} & 60\%$\to$\textbf{55\%} & 60\%$\to$\textbf{40\%} 
                            & \textbf{60\%} & 60\%$\to$\textbf{50\%} & 60\%$\to$\textbf{35\%} & 60\%$\to$\textbf{50\%} \\
            \bottomrule
        \end{tabular}%
    }
\end{table*}

\textbf{Fine-grained action prediction in dynamic interactions.} 
Our tasks challenges the model to handle rigid, articulated, and deformable objects, requiring precise low-level control under complex dynamics.
In the P\&P task, which necessitates simultaneous obstacle avoidance and precise end-effector placement, UniVLA falters due to the coarse resolution of its high-level latent planning. While VPP captures dynamics via video prediction, it is consistently outperformed by \model. This comparison underscores the advantage of explicitly modeling predictive information in a dedicated action condition space, rather than relying on high-dimensional visual predictions.
This advantage is further amplified in the \textit{Fold} task. Controlling deformable objects requires accurate trajectory planning and precise release timing to achieve the target geometry. \model significantly widens the performance gap over VPP, as its condition space effectively distills manipulation-relevant dynamics (e.g., cloth deformation) while discarding the redundant perceptual signals inherent in video generation. Similarly, in the \textit{Close} task, \model achieves near-perfect performance, confirming that the predicted conditions are sufficient to guide fine-grained interaction with articulated dynamics.

\textbf{Generalization ability in OOD scenes.} 
A key aspect of generalization from upstream pretraining to downstream finetuning is the ability to transfer manipulation knowledge across heterogeneous embodiments, while preserving reliance on high-level visual representations rather than overfitting to the finetuning environment.
Under background and novel object variations, baseline methods exhibit pronounced degradation. Latent action models tend to implicitly overfit to object-specific dynamics coupled with training appearances, limiting transferability. Similarly, VPP is constrained by the visual distribution of expert demonstrations, leading to artifacts when inputs deviate from the training domain.
In contrast, \model maintains superior performance with minimal degradation across all OOD scenarios. 
We attribute this robustness to the design of the condition space: by querying and compressing informative features from \textit{frozen}, pretrained visual encoders, \model constructs conditions that are highly distinctive for manipulation yet invariant to visual nuisances. 
This design ensures that the generalization power of upstream visual priors is preserved rather than distorted during fine-tuning. 
Notably, even under severe lighting changes, as the most challenging shift, \model exhibits the smallest relative performance drop. This confirms that our framework learns stable, action-centric representations, striking an optimal balance between expressiveness for control and compactness for generalization.

\subsection{Efficient Training Strategy}
We pretrain \model and its variants on the OXE dataset under different training strategies as ablations.
\textit{vanilla VLA} follows standard VLA pretraining with the same VLM backbone and DiT head, supervising actions from current observations only.
\textit{\model w/o cotrain} is trained using the first-stage future-observation-guided objective and second-stage action supervision without condition supervision.
All models are fine-tuned with identical expert demonstrations on downstream tasks. Details are shown in Appendix~\ref{app:ablation}.
\begin{table*}[h!]
    \centering
    \caption{\textbf{Ablation of training stages in \model.} \textit{Vanilla VLA} predicts actions solely from current observations. \textit{\model w/o cotrain} removes the condition supervision in the second stage. We report success rates (\%); for OOD settings, scores are formatted as \textit{ID $\to$ OOD}. Best results are highlighted in \textbf{bold}.}
    \label{tab:ablation}
    
    \definecolor{tabgray}{gray}{0.95}
    \setlength{\tabcolsep}{4pt}
    \renewcommand{\arraystretch}{1.12}
    
    \resizebox{0.95\textwidth}{!}{%
        \begin{tabular}{l | c | ccc | cccc}
            \toprule
            \multirow{2}{*}{\textbf{Model}} & \textbf{Microwave} & \multicolumn{3}{c|}{\textbf{Pick and Place}} & \multicolumn{4}{c}{\textbf{Fold the Towel}} \\
            \cmidrule(lr){2-2} \cmidrule(lr){3-5} \cmidrule(lr){6-9}
             & ID & ID & Background & Novel Object & ID & Background & Light Change & Novel Object \\
            \midrule
            
            Vanilla VLA 
                & 90\% 
                & 45\% & 45\%$\to$45\% & 45\%$\to$\textbf{40\%} 
                & 40\% & 40\%$\to$25\% & 40\%$\to$10\% & 40\%$\to$30\% \\
            
            \model~w/o cotrain 
                & 95\% 
                & 45\% & 45\%$\to$45\% & 45\%$\to$35\% 
                & 30\% & 30\%$\to$30\% & 30\%$\to$10\% & 30\%$\to$30\% \\
            
            \rowcolor{tabgray}
            \textbf{\model} (Ours) 
                & \textbf{100\%} 
                & \textbf{60\%} & 60\%$\to$\textbf{55\%} & 60\%$\to$\textbf{40\%} 
                & \textbf{60\%} & 60\%$\to$\textbf{50\%} & 60\%$\to$\textbf{35\%} & 60\%$\to$\textbf{50\%} \\
            \bottomrule
        \end{tabular}%
    }
\end{table*}

As shown in Table~\ref{tab:ablation}, \model trained with the two-stage scheme outperforms both variants.
Compared to the \textit{vanilla VLA}, \model achieves improvements across all tasks under the ID setup, demonstrating its ability to extract effective conditions from observations.
Moreover, under the OOD setup, \model maintains strong generalization performance, indicating the robustness of the predictive condition space.
For \textit{\model w/o cotrain}, we observe performance comparable to the \textit{vanilla VLA} across tasks. This suggests that introducing future conditions does not significantly degrade the action generation capability of the VLM backbone.
However, this variant still falls notably behind the full \model, highlighting the necessity of the co-training. These results confirm that explicitly supervising the alignment between future conditions and the VLM backbone is crucial for distilling future action-relevant knowledge into the VLA model.

\subsection{Learning from Human Data}
\label{sec:human_exp}
We evaluate the two human data utilization strategies introduced in Section~\ref{sec:humanumi}, with results summarized in Table~\ref{tab:human}.
\textit{w. human v.} denotes leveraging unannotated human videos only in the second stage for future condition supervision.
\textit{w. human v./a.} further incorporates a small subset of action-annotated human videos, which are supervised for action prediction in both training stages. Specific dataset information and training strategies are introduced in Appendix~\ref{app:human}.

\begin{table*}[h!]
    \centering
    \caption{\textbf{Performance under different human data integration strategies.} \textit{w/o human data}: trained solely on robot data. \textit{w. human v.}: trained with unannotated human videos. \textit{w. human v./a.}: trained with a mix of annotated and unannotated human videos. Best results are \textbf{bold}.}
    \label{tab:human}
    
    \definecolor{tabgray}{gray}{0.95}
    \setlength{\tabcolsep}{6pt} 
    \renewcommand{\arraystretch}{1.12}
    
    \resizebox{0.9\textwidth}{!}{%
        \begin{tabular}{l | ccc | cccc}
            \toprule
            \multirow{2}{*}{\textbf{Strategy}} & \multicolumn{3}{c|}{\textbf{Pick and Place}} & \multicolumn{4}{c}{\textbf{Fold the Towel}} \\
            \cmidrule(lr){2-4} \cmidrule(lr){5-8}
             & ID & Background & Novel Object & ID & Background & Light Change & Novel Object \\
            \midrule
            
            w/o human data (Base)
                & 60\% & 60\%$\to$55\% & 60\%$\to$40\% 
                & 60\% & 60\%$\to$50\% & 60\%$\to$35\% & 60\%$\to$\textbf{50\%} \\
            
            w. human v.
                & \textbf{70\%} & 70\%$\to$\textbf{70\%} & 70\%$\to$35\% 
                & 50\% & 50\%$\to$45\% & 50\%$\to$30\% & 50\%$\to$45\% \\
            
            \rowcolor{tabgray}
            \textbf{w. human v./a.}
                & \textbf{70\%} & 70\%$\to$\textbf{70\%} & 70\%$\to$\textbf{45\%} 
                & \textbf{65\%} & 65\%$\to$\textbf{60\%} & 65\%$\to$\textbf{45\%} & 65\%$\to$\textbf{50\%} \\
            \bottomrule
        \end{tabular}%
    }
\end{table*}

As shown in Table~\ref{tab:human}, we observe that even when human videos are used solely for condition prediction supervision in the second stage, our model still benefits on the P\&P task and exhibits a smaller relative performance drop under OOD settings.
However, performance degrades on the deformable object manipulation task.
We attribute this behavior to the task-dependent similarity between human and robotic manipulation.
For pick-and-place, task execution patterns in human demonstrations closely resemble robotic behaviors, leading to more aligned action-relevant conditions.
In contrast, deformable object manipulation induces a larger mismatch in the condition space for many conditions from more flexible human manipulations are not modeled during first-stage robot training, which limits the transferability of human manipulation knowledge and results in degraded performance.

Nevertheless, even with only 220h of action-annotated human data introduced in the first training stage, \model is able to rapidly acquire human-aligned conditioning representations and effectively transfer them to robotic manipulation.
The resulting model consistently outperforms its robot-only counterpart across all ID and OOD settings, demonstrating substantially improved generalization.
These results indicate the strong potential of our framework to scale with larger and more diverse human datasets.

\subsection{Learning from UMI Data}
\label{sec:umi_exp}

We collected an additional 120 UMI trajectories for both the P\&P and \textit{Fold} tasks to augment the training process detailed in Appendix~\ref{app:umi}. 
Crucially, this data was introduced exclusively during the second-stage fine-tuning alongside our expert demonstrations, ensuring that the condition space learned in the first stage remained unaltered.
Despite the significant domain gaps characterized by UMI's egocentric observations, distinct action representations, and a completely different embodiment configuration: the fact that the condition space was established solely on OXE pretraining without prior exposure to such inputs, \model achieved remarkable performance gains. 

\begin{wrapfigure}{r}{0.42\textwidth}
  \centering
    \includegraphics[width=0.4\textwidth]{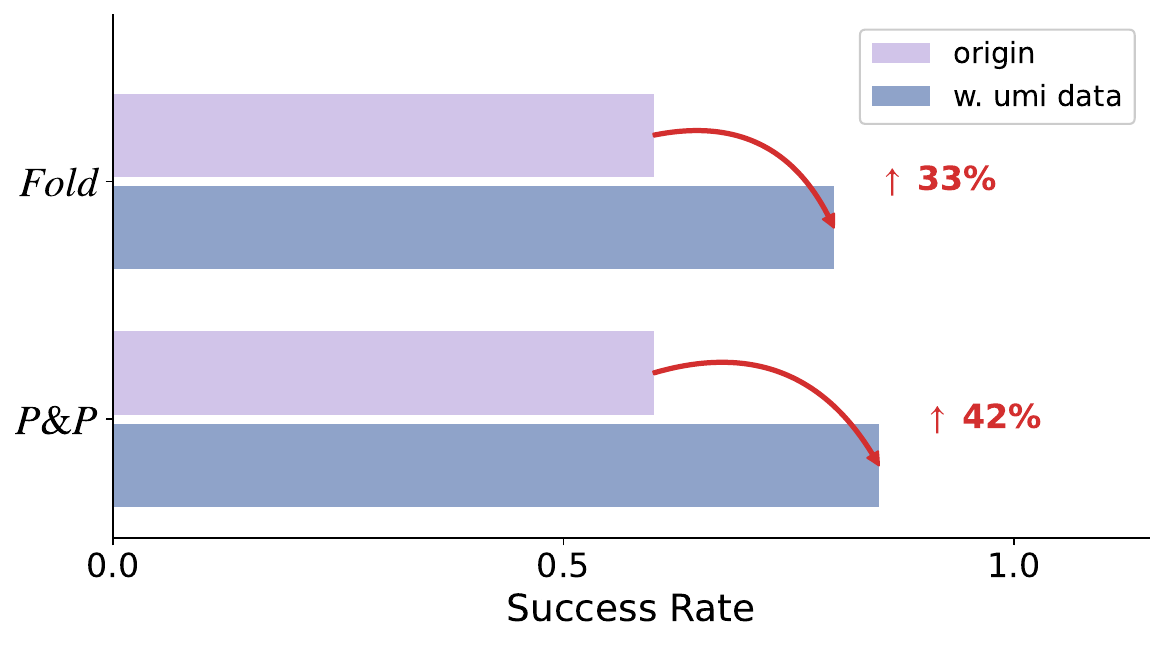}
    \caption{\textbf{Performance after training with UMI data.} Compared to training solely with robot data, the \model showed a 42\% improvement in performance on the P\&P task and a 33\% on the Fold task.}
\label{umiexp}
\end{wrapfigure}

Specifically, success rates surged from 60\% to 85\% on P\&P and from 60\% to 80\% on \textit{Fold} as shown in Figure~\ref{umiexp}. 
These results demonstrate that even when pretrained strictly on standard robot data, \model acquires a highly robust and generalizable condition encoding capability. 
We attribute this success to the model's proficiency in capturing embodiment-agnostic dynamics, such as intrinsic object motion, which facilitates the seamless integration of UMI data and highlights the broad scalability of our approach.
\section{Conclusion}
We propose \model, a novel world modeling framework for action generation that compresses future observations into a low-dimensional condition space via guidance.
By jointly predicting these compact future conditions within a VLA, \model strikes an effective balance between efficient future forecasting and the acquisition of rich manipulation knowledge.
Experiments demonstrate the efficiency and strong generalization capability of our approach, while validating the effectiveness of the proposed training strategy and the scalability of \model to large-scale human manipulation data.
Future work may focus on designing more expressive and efficient condition representations to better handle scenarios with strong spatial or action constraints, as well as exploring improved knowledge distillation and more generalizable condition learning from human videos.

\clearpage
\beginappendix
\section{Detailed Architecture}
\label{app:model}
Our method samples future visual observations at a temporal frequency of one-quarter relative to the action sequence. 
Concretely, for the default prediction horizon of 16 action steps, we uniformly sample 4 frames to serve as the future observation sequence. 
Regarding feature encoding, the frozen DINOv2 model processes each sampled image to extract semantic features. 
In parallel, for the Wan VAE, we utilize the current observation as the initiating frame, encoding it jointly with the sampled future frames to capture temporal and spatial features. 
Both the DINOv2 features and the VAE features (with their last two spatial dimensions flattened) are projected into a unified embedding space matching the hidden dimension of the Q-Former and are subsequently stacked. 
A Q-Former, instantiated with $N=16$ learnable query tokens, aggregates the necessary dynamic representations from this combined feature set via cross-attention mechanisms, ultimately compressing the information into a compact condition space with a dimensionality of $D=32$ by default.

The query mechanism of the Q-Former-based Encoder remains consistent across both training stages, with the sole distinction being that the encoder transitions from a trainable state in the first stage to a frozen state in the second.
In this second stage, the VLM processes the current observation to yield the sequence of last hidden states. 
Capitalizing on the causal nature of the architecture, the trailing tokens effectively distill the most comprehensive visual and linguistic context—analogous to our strategy of utilizing the final learnable token as the action token. 
Consequently, we select the last 4 tokens from the VLM output to align with the target future representations extracted by the frozen vision foundation models.
To implement this alignment, we instantiate 16 learnable query embeddings, matching the token count of the condition representation. 
These embeddings perform cross-attention over the selected VLM hidden states and are subsequently projected into the 32-dimensional space to compute the alignment loss against the ground-truth future conditions.

\begin{figure*}[h!]
  \centering
    \includegraphics[width=\textwidth]{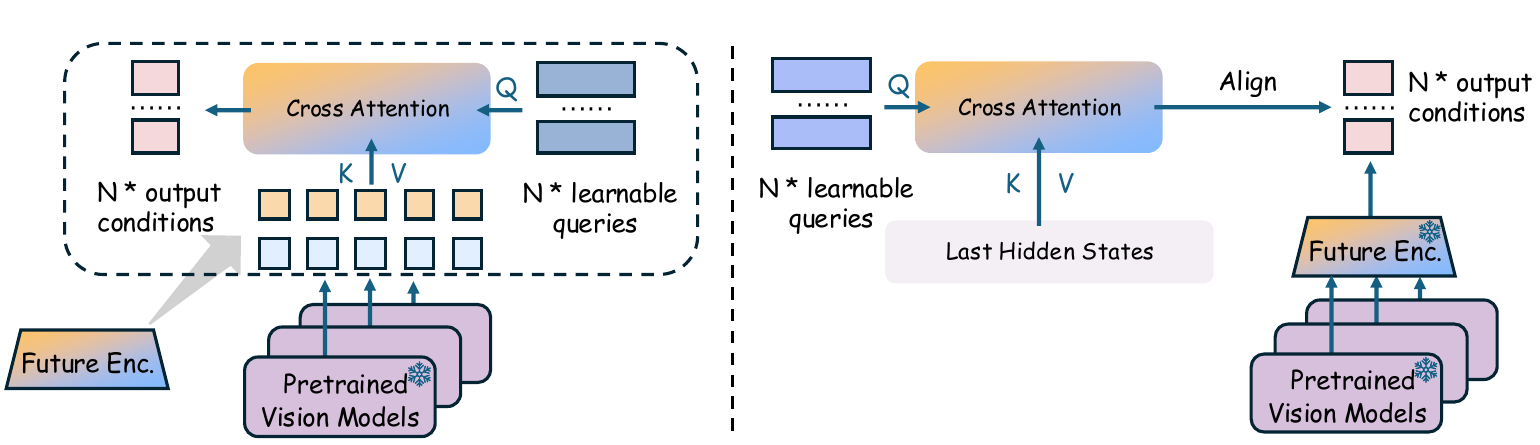}
    \caption{\textbf{Detailed illustration of the query mechanisms within \model.} 
The \textbf{left panel} depicts the future encoder, which maintains a set of learnable queries to extract low-dimensional, action-relevant conditions from the features of pretrained vision models. 
The \textbf{right panel} illustrates the query mechanism for condition prediction during the second training stage. 
Here, a set of learnable query embeddings performs cross-attention with the last hidden states of the VLM, producing predictive representations that are supervised to align with the target conditions generated by the now-frozen future encoder.}
\label{app:detail}
\end{figure*}

We detailed above query mechanisms in Figure~\ref{app:detail}.

\section{Simulation Experiments}
\label{app:sim}
\subsection{Training Settings}
For simulation evaluation, Our model is pretrained on the Open X-Embodiment (OXE) dataset~\cite{oxe} for 100k training steps with a global batch size of 1024 in the first stage, the dataset sampling ratios are shown in Table~\ref{tab:oxe_ratio}.
In the second stage, the model is further trained on the Bridge and Fractal datasets for 50k steps using the same batch size with individual ratios as about 51\% and 49\%.
No additional fine-tuning is needed on individual Bridge or Fractal datasets.
\begin{table*}[h!]
\centering
\caption{Dataset Sampling Ratios during the OXE pretraining.}
\label{tab:oxe_ratio}
\resizebox{\textwidth}{!}{
\begin{tabular}{l c @{\hspace{1.2em}} l c @{\hspace{1.2em}} l c}
\toprule
Dataset & Ratio & Dataset & Ratio & Dataset & Ratio \\
\midrule
Fractal & 0.140639 &
Kuka & 0.140674 &
Bridge & 0.146649 \\

Taco Play & 0.032661 &
Jaco Play & 0.005354 &
Berkeley Cable Routing & 0.002907 \\

Roboturk & 0.025753 &
Viola & 0.010483 &
Berkeley Autolab UR5 & 0.013452 \\

Toto & 0.022367 &
Stanford Hydra Dataset & 0.049202 &
Austin Buds Dataset & 0.002343 \\

NYU Franka Play Dataset & 0.009245 &
Furniture Bench Dataset & 0.027112 &
UCSD Kitchen Dataset & 0.000545 \\

Austin Sailor Dataset & 0.024248 &
Austin Sirius Dataset & 0.019224 &
DLR EDAN Shared Control & 0.000613 \\

IAMLab CMU Pickup Insert & 0.010043 &
UTAustin Mutex & 0.024852 &
Berkeley Fanuc Manipulation & 0.008600 \\

CMU Stretch & 0.001718 &
BC-Z & 0.082621 &
FMB Dataset & 0.023224 \\

DobbE & 0.015656 &
DROID & 0.111433 &
 &  \\
\bottomrule
\end{tabular}
}
\end{table*}

\subsection{Evaluation Settings}
The SIMPLER simulator~\cite{simplerenv} distinguishes itself by functioning as a rigorous real-to-sim evaluation proxy for the Fractal and Bridge datasets, thereby compelling models to tackle cross-domain transfer challenges. 
Notably, the simulator offers two evaluation protocols characterized by distinct degrees of domain shift for Google Robot tasks: 
\textit{Visual Matching}, which aims to closely replicate real-world tasks by minimizing visual discrepancies between simulated and physical environments; 
and \textit{Variant Aggregation}, which introduces more severe distributional shifts by modifying environmental elements such as backgrounds, lighting conditions, distractors, and table textures.

\section{Real-World Experiments}
\label{app:real}
\subsection{Training Settings}
For real-world experiments, our model is pretrained on the Open X-Embodiment (OXE)
dataset for 100k training steps with a global batch size of
1024 in the first stage and 50k steps on OXE in the second stage. We also use the OXE pretrained checkpoints provided by the baselines. All the models are then finetuned on our real-world episodes for 30 epochs with a global batch size of 512 (During the fine-tuning phase, \model mirrors the training protocol of only the second stage, maintaining the co-training objective that jointly optimizes for both action prediction and future condition forecasting). The oxe pretraining dataset sampling ratios are shared with Table~\ref{tab:oxe_ratio}. For our self-collected finetuning dataset, all the tasks share equal sampling weights, Therefore, the sampling probability is only positively correlated with the training sample amount for each task.
\subsection{Evaluation Settings}
 We adopt Maniunicon~\cite{maniunicon} control strategy with the same evaluation settings. For all methods, the policy predicts a future action sequence of length 16 conditioned on the current RGB observation from the D435 camera, of which the first 8 steps are executed. As mentioned above, we continue to use the default DINOv2 and Wan VAE as the pre-trained visual encoder combination for our real-world experiments.
\subsection{Ablation Details}
\label{app:ablation}
For the \textit{Vanilla VLA}, we adhere to a standard training protocol where the model is supervised solely on action prediction given current observations. 
To ensure a fair comparison, we align the training budget with the total duration of the two-stage \model: the model is pretrained on the OXE dataset for 150k steps with a global batch size of 1024, followed by fine-tuning on expert demonstrations using a batch size of 512 for 30 epochs.

Regarding the \textit{\model w/o cotrain} variant, the first stage setup remains identical to that of the full \model, involving 100k steps of pretraining with a batch size of 1024 under future-observation-guided action supervision. 
In the second stage, the model continues training for an additional 50k steps with the same batch size. 
However, crucially, we exclude the co-training objective in this phase; the model receives supervision exclusively for action prediction, without the auxiliary supervision for predicting future conditions. Subsequently, the model undergoes fine-tuning on expert demonstrations for 30 epochs with a global batch size of 512. 
Consistent with the second pretraining stage, supervision during this fine-tuning phase remains restricted exclusively to action prediction.

\subsection{Human Data Learning Settings}
\label{app:human}
\begin{figure}[h!]
  \centering
    \includegraphics[width=0.9\textwidth]{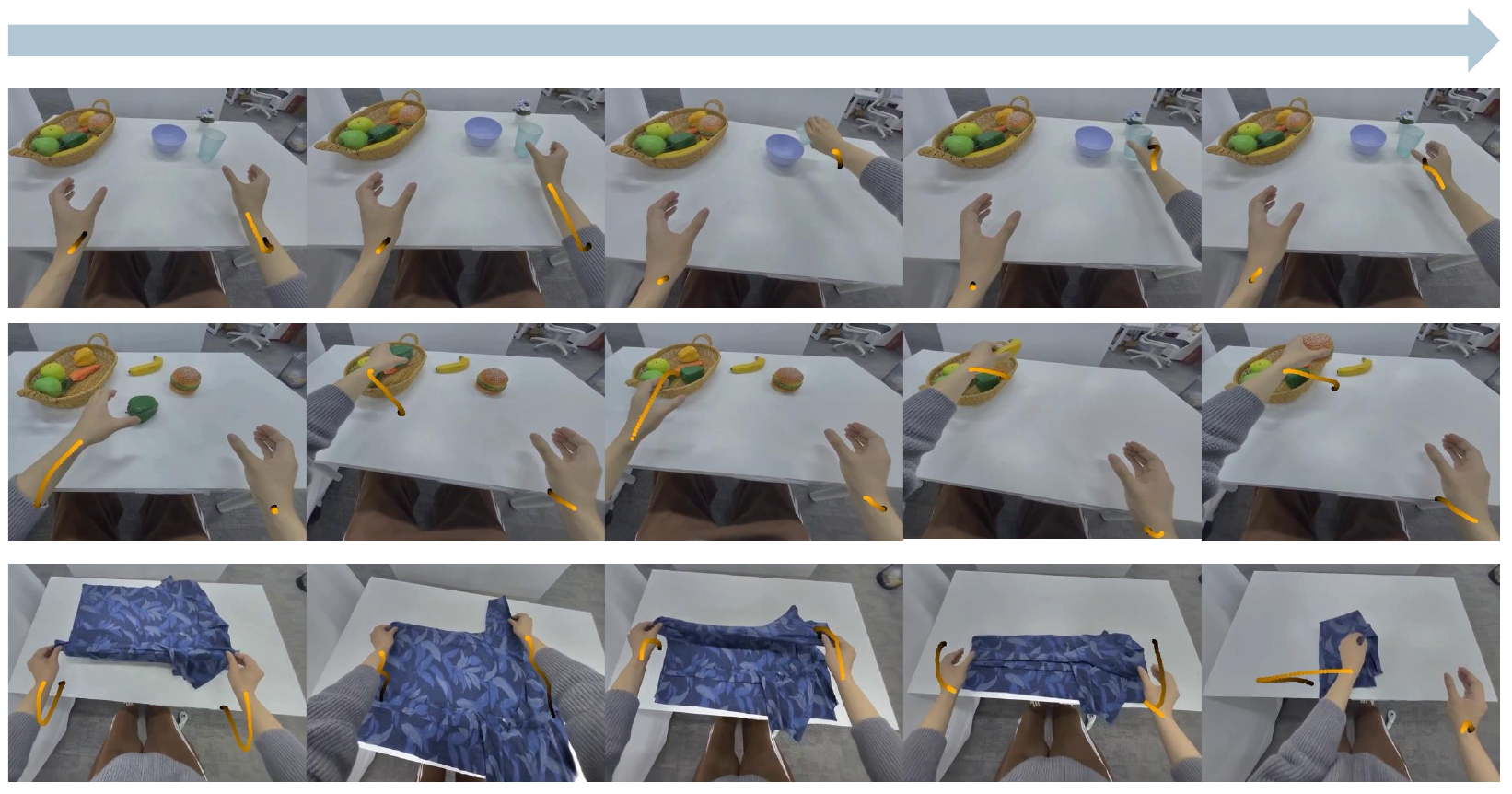}
    \caption{\textbf{Process of human manipulation process.} We keep the same human manipulation data collection strategy as in~\cite{gr3,gr-dexter}. Human trajectories can be efficiently collected with
VR devices at a rate of approximately 450 trajectories per hour, substantially outpacing the teleoperated
robot trajectory collection. Nevertheless, as discussed earlier, we use action annotations for only 11\% of the data, while the remaining 89\% are leveraged solely as unlabeled videos.
We argue that this setting more faithfully reflects real-world conditions, where action-labeled human manipulation data are scarce but large-scale unlabeled videos are abundant, and is therefore essential for evaluating the scalability of \model.
}
\label{pico}
\end{figure}
\textbf{Dataset Information.} We use a human manipulation dataset collected by ourselves with PICO 4 Ultra Enterprise as in~\cite{gr3}, consisting of \textbf{650k} trajectories with a total duration of approximately \textbf{1,920} hours, among which a \textbf{220}-hour subset is annotated with actions.
The human arm state is parameterized by the pose (translation and rotation) of the wrist joint relative to the PICO coordinate system. Additionally, the hand aperture (palm opening and closing) is mapped to the continuous/discrete state of the robot gripper. To facilitate a unified action representation, we formulate actions as the state deltas between consecutive frames, expressed in the reference frame of the preceding time step.

\begin{table*}[h!]
\centering
\caption{Dataset Sampling Ratios for the Human Data Learning.}
\label{tab:oxe_huamn_ratio}
\resizebox{\textwidth}{!}{
\begin{tabular}{l c @{\hspace{1.2em}} l c @{\hspace{1.2em}} l c}
\toprule
Dataset & Ratio & Dataset & Ratio & Dataset & Ratio \\
\midrule
Fractal & 0.117016 &
Kuka & 0.117045 &
Bridge & 0.122016 \\

Taco Play & 0.027175 &
Jaco Play & 0.004455 &
Berkeley Cable Routing & 0.002419 \\

Roboturk & 0.021427 &
Viola & 0.008722 &
Berkeley Autolab UR5 & 0.011192 \\

Toto & 0.018610 &
Stanford Hydra Dataset & 0.040937 &
Austin Buds Dataset & 0.001949 \\

NYU Franka Play Dataset & 0.007692 &
Furniture Bench Dataset & 0.022558 &
UCSD Kitchen Dataset & 0.000454 \\

Austin Sailor Dataset & 0.020175 &
Austin Sirius Dataset & 0.015995 &
DLR EDAN Shared Control & 0.000510 \\

IAMLab CMU Pickup Insert & 0.008356 &
UTAustin Mutex & 0.020677 &
Berkeley Fanuc Manipulation & 0.007155 \\

CMU Stretch & 0.001429 &
BC-Z & 0.068743 &
FMB Dataset & 0.019323 \\

DobbE & 0.013026 &
DROID & 0.092715 &
Human Manip. Data & 0.167972 \\
\bottomrule
\end{tabular}
}
\end{table*}

\textbf{Training Strategy.} We explain the specific implementation of the two methods in section~\ref{sec:humanumi} in our experiment.

For the \textit{w. human v.} variant: \model 
is pretrained on the Open X-Embodiment (OXE)
dataset for 100k training steps with a global batch size of
1024. At the second stage, all 1,920 hours of unannotated human videos are used with the OXE dataset. For robot data, the model receives supervision from both condition prediction and action prediction. For human video data, it receives supervision only from condition prediction.
50k steps of the same batch size is also used in the second stage.

For \textit{w. human v./a.} variant: action supervision is applied to the 220-hour annotated subset in both training stages. 
Specifically, in the first stage, the model utilizes both the OXE dataset and the 220-hour annotated human subset to perform action prediction conditioned on future observations.
In the second stage, the training set is expanded to include the full 1,920-hour corpus of human manipulation videos alongside the OXE dataset.
During this phase, action supervision is applied exclusively to the OXE data and the 220-hour annotated human subset, while the supervision for future condition prediction is applied to all data samples. 

The dataset sampling ratios incorporated human dataset are shown in Table~\ref{tab:oxe_huamn_ratio}. We also show our human manipulation video samples in Figure~\ref{pico}.

\subsection{UMI Data Learning Settings}
\label{app:umi}
Our UMI data collection consists exclusively of egocentric observations, where actions are defined as egocentric motions relative to the headset coordinate system.
Since our objective is to evaluate the performance improvement on our target tasks, we incorporate UMI data solely as a data source during the final fine-tuning stage alongside expert demonstrations.
Consistent with the sampling strategy described earlier, it is assigned a sampling weight equivalent to that of the other expert demonstration tasks.
\clearpage
\bibliographystyle{plainnat}
\bibliography{ref}
\end{document}